\documentclass[letterpaper]{article} 
\usepackage{aaai2026}  
\usepackage{times}  
\usepackage{helvet}  
\usepackage{courier}  
\usepackage[hyphens]{url}  
\usepackage{graphicx} 
\usepackage{natbib}  
\usepackage{caption} 
\usepackage{algorithm}
\usepackage{algorithmic}

\usepackage{amsmath}
\usepackage{multirow}
\usepackage{makecell}

\usepackage{newfloat}
\usepackage{listings}
\DeclareCaptionStyle{ruled}{labelfont=normalfont,labelsep=colon,strut=off} 
\floatstyle{ruled}
\newfloat{listing}{tb}{lst}{}
\floatname{listing}{Listing}
\nocopyright

\title{Adapting AlignScore Mertic for Factual Consistency \\ Evaluation of Text in Russian: A Student Abstract}
\author {
    Mikhail Zimin,
    Milyausha Shamsutdinova,
    Georgii Andriushchenko
}
\affiliations {
    Innopolis University\\
    \{m.zimin, m.shamsutdinova, g.andryushchenko\}@innopolis.university
}

\usepackage{bibentry}

\begin{document}

\maketitle

\begin{abstract}
Ensuring factual consistency in generated text is crucial for reliable natural language processing applications. However, there is a lack of evaluation tools for factual consistency in Russian texts, as existing tools primarily focus on English corpora. To bridge this gap, we introduce \textsc{AlignRuScore}, a comprehensive adaptation of the AlignScore metric for Russian. To adapt the metric, we fine-tuned a RuBERT-based alignment model with task-specific classification and regression heads on Russian and translated English datasets. Our results demonstrate that a unified alignment metric can be successfully ported to Russian, laying the groundwork for robust multilingual factual consistency evaluation. We release the translated corpora, model checkpoints, and code to support further research.
\end{abstract}

\begin{links}
    \textbf{Code} \\ 
    https://github.com/MilyaushaShamsutdinova/AlignRuScore \\
    \textbf{Data} \\
    https://huggingface.co/collections/MilyaShams/alignscore-russian-datasets-6801082af31d0a240e4b9bb5 \\
    \textbf{Model} \\
    https://huggingface.co/CatFr0g/ruAlignScore
\end{links}

\section{Introduction}

Modern natural language generation systems are increasingly used in critical applications where factual correctness is imperative. AlignScore \cite{zha2023alignscoreevaluatingfactualconsistency} introduced a unified alignment function that maps pairs of text (context and claim) to a factual consistency score. 
Since AlignScore was originally designed for English texts,
applying AlignScore directly to Russian texts may be invalid. This paper proposes a strategy to adapt the AlignScore metric to Russian texts to increase its reliability when evaluating the factual consistency of text generated in Russian.

\section{Related Work}

AlignScore has emerged as a powerful metric for measuring factual consistency by unifying diverse natural language processing (NLP) tasks. See Supplementary Materials for AlignScore details. QAFactEval \cite{fabbri-etal-2022-qafacteval} and UniEval \cite{li2025unieval} metrics have shown promising results but remain primarily focused on English data. As for the fact alignment datasets, there are two such datasets in Russian. Russian Semantic Text Similarity (RuSTS) \cite{ru_sts} is a human translation of Semantic Text Similarity Benchmark (STSB) \cite{cer-etal-2017-semeval}. RuFact\footnote{\url{https://kaggle.com/competitions/internal-fact-checking-for-the-russian-language}} is a paraphrase dataset created with various approaches for the generation of training examples.

\section{Methodology}

Our methodology adapts the original AlignScore approach to the Russian language, resulting in AlignRuScore. The process involved data collection and translation, unified model training, adaptation of the metric's calculation strategy, and evaluation of the resulting metric.

\subsection{Data Collection and Translation} 

We constructed a diverse Russian training corpus by translating subsets (up to 10,000 examples each, where applicable) of the English datasets used in the original AlignScore paper, covering tasks such as Natural Language Inference (NLI), Fact Verification, Paraphrasing, Question Answering (QA), and Semantic Textual Similarity (STS). Machine translation was performed using Yandex Translate\footnote{\url{https://translate.yandex.com}}. The translated datasets were supplemented with native Russian datasets such as RuFacts for paraphrase and fact verification and RuSTS benchmark. The final unified corpus comprised over 118,000 training examples spanning multiple alignment-related tasks, as shown in Table 5 of the Supplementary Materials.

\subsection{Adapting AlignScore Metric to Russian}

\subsubsection{Training Unified Alignment Function on Russian Corpus}

Following the original methodology of AlignScore, we employed a RuBERT-base model \cite{rubert} as the backbone for our alignment function. The model was fine-tuned using a unified multi-task learning approach on the collected Russian corpus. It features task-specific output heads for 1) 3-way classification (predicting entailment, neutral, or contradiction); 2) binary classification (predicting whether context and claim are aligned or not-aligned); 3) regression (predicting a similarity score between 0 and 1).
 
The model was trained by optimizing a combined loss function encompassing cross-entropy for classification heads and mean squared error for the regression head, allowing it to learn a generalized representation of textual alignment from the diverse data sources. Training hyperparameters are detailed in Table 1 of the Supplementary Materials.

Following AlignScore’s original formulation, we train a small feed-forward network for each type of training task (see Table 5 of Supplementary Materials for the training tasks) and learn them simultaneously, so the embedding space and the heads can properly handle alignment for all NLP tasks.

\subsection{Evaluation on Russian datasets}

We evaluated AlignRuScore on held-out test portions of Russian datasets corresponding to the core alignment tasks on which we trained the metric. Performance was measured using standard metrics appropriate for each task: Accuracy, Precision, Recall, F1, and ROC AUC for classifications; $R^2$ and Mean Squared Error (MSE) for regression.

\section{Evaluation Results}

\subsection{Evaluation of Resulting Metric}

Detailed evaluation results are presented in Tables 2-4, reporting results across different types of tasks and metrics.

\subsection{Evaluating LLMs using Resulting Metric}

To further validate the applicability of AlignRuScore, we conducted an evaluation of the Gemini 1.5-Flash\footnote{\url{https://deepmind.google/models/gemini/flash/}} model on a held-out subset of 200 examples. The model’s outputs were scored using our adapted metric, yielding a mean factual-consistency score of 
$0.7285 \pm 0.0639.$ 

\subsection{Discussion of Results}

The evaluation results demonstrate strong results of AlignRuScore on NLI, fact verification, and entailment detection tasks. In paraphrase detection, the model shows a robust behavior on well-structured question pairs.

However, claim verification under terse or noisy contexts remains an issue. In question answering, there is a tendency to over-generate candidate answers. In information retrieval, precision is improved, but still remains below paraphrase and inference levels. It highlights the need for more precise answer extraction.

\section{Conclusion}

In this work, we presented \textsc{AlignRuScore}, a Russian-language adaptation of the unified factual consistency metric AlignScore. By translating and augmenting a diverse set of English benchmarks—spanning natural language inference, fact verification, paraphrase detection, semantic textual similarity, question answering, and information retrieval—and fine-tuning a RuBERT-based alignment model, we obtained a single alignment function capable of handling the nuances of Russian morphology and syntax.

\section{Future Work}

Looking ahead, we plan to (1) incorporate additional Russian-native datasets—particularly for summarization and dialogue consistency; (2) explore architecture variants, such as multilingual transformer backbones and task-adaptive adapters; and (3) evaluate \textsc{AlignRuScore} in downstream applications, including automated fact-checking and evaluation of Russian-language generative models. We release our code, translated corpora, and trained checkpoints to facilitate further research in Russian NLP.

\bibliography{aaai2026}

\newpage

\section{Supplementary Materials}

\subsection{Factual Consistency Evaluation with AlignRuScore}

To evaluate the factual consistency between a given Russian context and claim, AlignRuScore utilizes the trained alignment function combined with the context/claim splitting and aggregation strategy using the following steps:

\begin{enumerate}

    \item The potentially long context is segmented into overlapping chunks of approximately 350 tokens at sentence boundaries.
    
    \item The claim is split into individual sentences.
    
    \item Each claim sentence is evaluated against every context chunk using the alignment model's 3-way classification head. The probability assigned to the 'aligned' (entailment) class is used as the alignment score.
    
    \item For each claim sentence, the maximum alignment score across all context chunks is selected, representing the strongest support found in the context for that specific sentence.
    
    \item The final AlignRuScore is the average of these maximum scores across all sentences in the claim.

\end{enumerate}

This chunking and aggregation approach allows the metric to handle long contexts effectively while producing a fine-grained score sensitive to inconsistencies at the sentence level within the claim.

\subsection{Datasets}

The detailed description and statistics of the data used for training and testing are provided in Table \ref{table:datasets}. 

\subsection{Alignment Model Training Hyperparameters}

\begin{table}[h]
\centering
\renewcommand{\arraystretch}{1}
\begin{tabular}{lcc}
\hline
\textbf{Hyperparameter} & \textbf{AlignRuScore-base} \\
\hline
Base Model      & RuBERT-base   \\
Parameters      & 180M           \\
Batch Size      & 12             \\
Epochs          & 3              \\
Optimizer       & AdamW           \\
Learning Rate   & 1e-5           \\
Weight Decay    & 0.1            \\
Adam $\epsilon$ & 1e-6           \\
Warmup Ratio    & 0.06          \\
Random Seed     & 2025           \\
GPU             & A100   \\
GPU Memory      & 20 GB \\
Time (hours)        & 4           \\
\hline
\end{tabular}
\caption{The hyperparameters used to train the alignment model.}
\label{table:hyperparameters}
\end{table}

\newpage

\subsection{Evaluation Results}

\begin{table}[h!]
  \centering
  
  \label{tab:3-way}
  \begin{tabular}{llrrrrr}
    \hline
    \textbf{Dataset} & \textbf{Precision} & \textbf{Recall} & \textbf{micro F1} & \textbf{Accuracy}  \\
    \hline
    SNLI     & 0.743 & 0.743 & 0.743 & 0.743 \\
    MultiNLI & 0.674 & 0.674 & 0.674 & 0.674 \\
    ANLI     & 0.692 & 0.692 & 0.692 & 0.692 \\
    FEVER    & 0.814 & 0.814 & 0.814 & 0.814 \\
    Vitamin C & 0.596 & 0.596 & 0.596 & 0.596 \\
    \hline
  \end{tabular}
  \caption{Results on 3-way classification}
\end{table}

\begin{table}[h!]
  \centering
  \label{tab:binary}
  \begin{tabular}{llrrrrrl}
    \hline
    \textbf{Dataset}     & \textbf{Precision} & \textbf{Recall} & \textbf{F1} & \textbf{ROC AUC} \\
    \hline
    RuFact      & 0.591 & 0.908 & 0.716 & 0.732 \\
    QQP         & 0.707 & 0.841 & 0.768 & 0.879 \\
    PAWS        & 0.613 & 0.896 & 0.728 & 0.766 \\
    DocNLI      & 0.762 & 0.954 & 0.847 & 0.897 \\
    RACE        & 0.250 & 0.998 & 0.400 & 0.524 \\
    MS MARCO   & 0.201 & 0.520 & 0.289 & 0.703 \\
    \hline
  \end{tabular}
  \caption{Results on binary classification}
\end{table}

\begin{table}[h!]
  \centering
  \label{tab:regression}
  \begin{tabular}{llrr}
    \hline
    \textbf{Dataset} & \textbf{MSE} & \boldmath$R^2$ \\
    \hline
    SICK     & 0.228 & 0.857 \\
    RuSTS  & 0.954 & 0.557 \\
    \hline
  \end{tabular}
  \caption{Results on regression}
\end{table}

\begin{table*}[!t]
\centering
\small
\begin{tabular}{@{}lllc@{}}
\hline
\textbf{Natural Lanugage Processing Task} & \textbf{Dataset} & \textbf{Training Task} & \textbf{Size}, $10^3$ \\
\hline
\multirow{4}{*}{\textit{Natural Language Inference}} 
    & SNLI \cite{bowman-etal-2015-large} & 3-way classification & 10 \\
    & MultiNLI \cite{williams-etal-2018-broad}  & 3-way classification & 10  \\
    & Adversarial NLI \cite{nie-etal-2020-adversarial} \ & 3-way classification & 10 \\
    & DocNLI \cite{yin-etal-2021-docnli}  & binary classification & 10 \\
\hline
\multirow{2}{*}{\textit{Fact Verification}} 
    & NLI-style FEVER \cite{nie_fever} & 3-way classification & 10 \\
    & Vitamin C \cite{schuster-etal-2021-get} & 3-way classification & 10 \\
\hline
\multirow{3}{*}{\textit{Paraphrase}} 
    & QQP \cite{quora_question_pairs_2017} & binary classification & 10 \\
    & PAWS \cite{zhang-etal-2019-paws} & binary classification & 10 \\
    & RuFacts* \cite{akozlova_rufacts}  & binary classification & 6.7 \\
\hline
\multirow{2}{*}{\textit{Semantic Text Similarity}} 
    & SICK \cite{marelli2014semeval} & regression & 4.4 \\
    & RuSTS* \cite{ru_sts} & regression & 7.8 \\
\hline
\textit{Question Answering} 
    & RACE \cite{lai-etal-2017-race} & binary classification & 10 \\
\hline
\textit{Information Retrieval} 
    & MS MARCO \cite{nguyen2016ms} & binary classification & 10 \\
\hline
    & & \textbf{Total} & \textit{118.9} \\
\end{tabular}
\caption{The training datasets of our alignment model. Datasets marked with a * were already translated or created in Russian. The other datasets were translated as a part of our work.}
\label{table:datasets}
\end{table*}

\end{document}